\documentclass[fleqn,10pt,twocolumn]{SICE_FES26}

\title{MarineCraft: Enabling Rapid Prototyping of Underwater Robots\\
via Modular Construction}

\author{Yuta Sugiura${}^{1\dagger}$}
\speaker{Yuta Sugiura}

\affils{${}^{1}$Department of Information and Computer Science, Keio University, Yokohama, Japan\\
(Tel: +81-45-566-1769; E-mail: sugiura@keio.jp)\\
}
\abstract{%
Underwater robot development is often hindered by the complexities of waterproofing and wiring, which significantly delay the rapid prototyping process. This paper presents MarineCraft, a modular toolkit designed to accelerate the development cycle through structural reconfiguration. The system features self-contained, waterproof propulsion modules that integrate power, wireless communication, and actuation. By eliminating centralized wiring and the need for repeated sealing, MarineCraft allows diverse robot geometries to be assembled and tested in minutes rather than days. Experimental results demonstrate that this reconfigurable architecture enables fast, iterative design cycles while maintaining reliable operation and leak-free performance at depths of up to 2.5 meters. Our toolkit effectively lowers the barrier to underwater robotics by transforming modularity into a vehicle for rapid physical prototyping.}


\begin{document}

\maketitle

\section{Introduction}
Underwater robots are increasingly utilized in disaster 
response, environmental monitoring, and scientific 
exploration. As their societal impact expands, accessible 
educational platforms for underwater robotics are becoming 
increasingly important.

However, constructing an underwater robot remains 
challenging for beginners. Waterproofing, structural assembly, 
and reliable actuation in water introduce significant practical 
barriers. As a result, most educational activities focus on 
operation rather than construction, limiting opportunities for 
hands-on prototyping experiences.

\begin{figure}[t]
  \centering
  \includegraphics[width=\columnwidth]{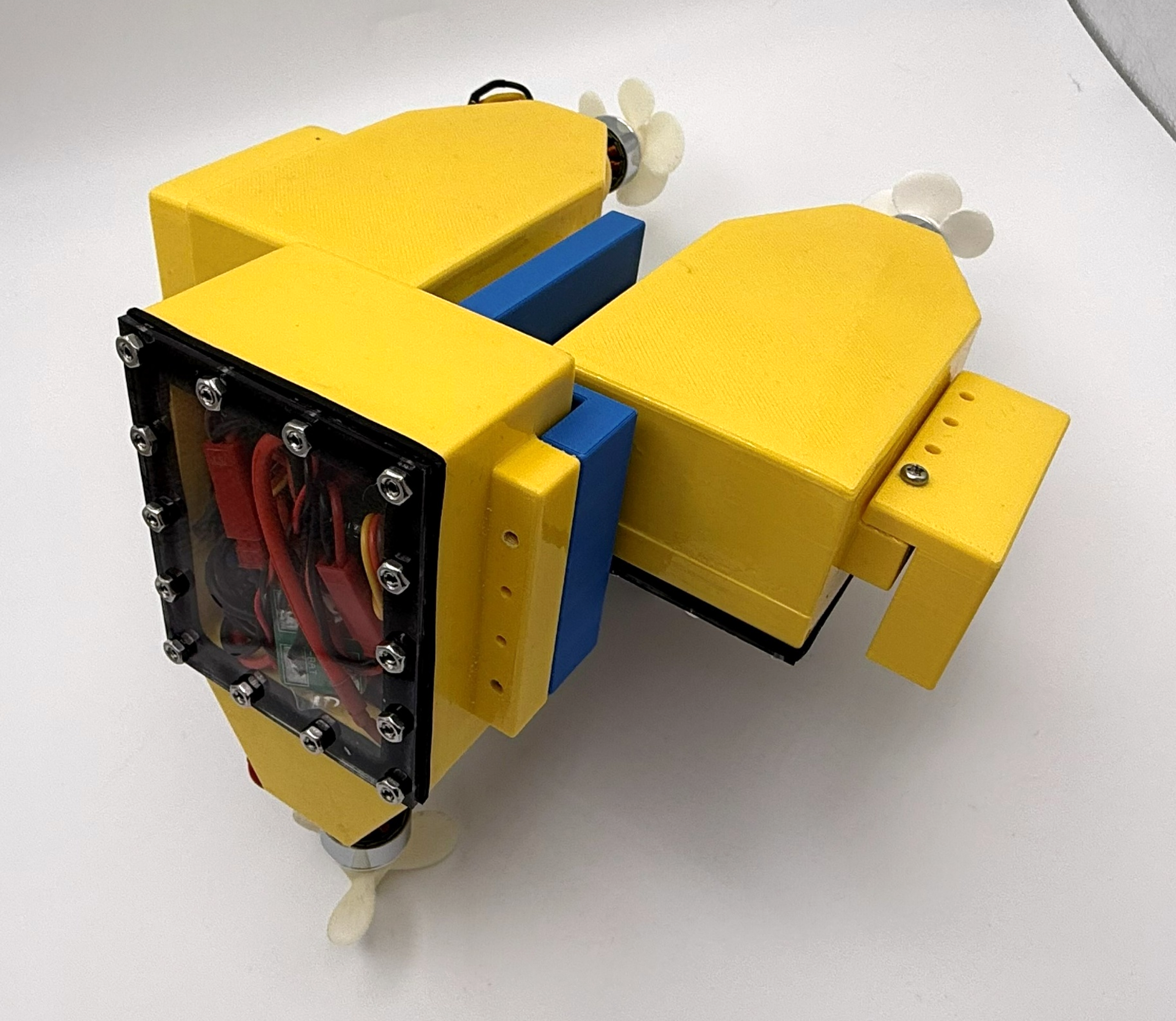}
  \caption{Overview of MarineCraft. The system employs self-contained modules to enable rapid physical prototyping of underwater robots through mechanical reconfiguration, eliminating the need for manual wiring and repeated waterproof sealing.}
  \label{fig:fig1}
\end{figure}

To address this gap, we propose MarineCraft, a modular 
underwater robot construction toolkit designed for hands-on 
educational use (Fig.~\ref{fig:fig1}). The system consists of independent 
waterproof motor modules, each integrating power, wireless 
communication, and actuation into a self-contained unit. 
By combining these modules with attachable structural 
components, users can assemble custom underwater robot 
configurations without requiring complex wiring or 
specialized fabrication skills.

Through system implementation and exploratory evaluation, 
we demonstrate that MarineCraft enables diverse underwater 
robot designs and lowers the barrier to underwater robot 
prototyping in educational contexts.

\section{Related Work}

\subsection{Underwater Robotics Platforms}
Underwater robots have been widely studied for applications such as deep-sea exploration, environmental monitoring, and infrastructure inspection. Research in this area has primarily focused on autonomous navigation, hydrodynamic modeling, and robust operation in complex submerged environments. 

Various locomotion mechanisms for underwater robots have been explored, including hybrid microrobotic designs employing specialized actuators for swimming control~\cite{MicroBiped}.

Yoerger et al.~\cite{Yoerger} demonstrated the at-sea deployment of an autonomous underwater vehicle (AUV) for deep-sea near-bottom surveys, integrating long-baseline acoustic positioning and terrain-following control algorithms.

In addition to research-grade systems, open-source and low-cost underwater platforms such as BlueROV2\footnote{Blue Robotics, ``BlueROV2 Underwater ROV,'' https://bluerobotics.com/store/rov/bluerov2/, accessed Mar. 1, 2026.} have broadened access to underwater robotics research by providing integrated hardware and control architectures. Santos et al.~\cite{roboticFish} presented a low-cost open-source robotic fish fabricated using 3D printing technology, demonstrating accessible underwater robotic prototyping.

Cie\'slak~\cite{stonefish} introduced Stonefish, an open-source marine robotics simulation tool with realistic hydrodynamics and ROS integration for testing underwater control and perception algorithms.

While these systems emphasize navigation performance and deployment reliability, structural reconfiguration and modular physical construction have received comparatively less attention. MarineCraft is a modular toolkit in which each module operates independently, eliminating the need for wiring management or cable length adjustment and enabling flexible structural configurations.

\subsection{Modular Robotics Systems}

Modular robotics has been extensively explored to enable structural reconfiguration and adaptive morphology. 
PolyBot~\cite{PolyBot} introduced a chain-type modular robotic system capable of reconfiguration through distributed actuation, demonstrating early feasibility of modular hardware architectures. 
M-TRAN~\cite{M-TRAN} further advanced self-reconfigurable robotic systems by integrating symmetric docking mechanisms and coordinated motion planning for shape transformation. 
Topobo~\cite{Topobo} explored constructive modular assembly with kinetic memory, emphasizing tangible interaction and learning through physically reconfigurable components. 
PINOKY~\cite{Pinoky} demonstrated externally attachable modular actuation that enables motion without intrusive structural modification, highlighting accessibility and ease of deployment.

While these systems established important foundations for modular robotic design, they are primarily developed for terrestrial environments and do not address waterproof integration or underwater operation. 
In contrast, MarineCraft extends modular construction principles to underwater contexts by integrating waterproofing, power, and wireless communication within self-contained propulsion units, enabling structural reconfiguration while maintaining functional underwater reliability.

\section{MarineCraft}

MarineCraft is a modular underwater robot construction system composed of independent propulsion units. Each propulsion function is implemented as a self-contained module, allowing users to assemble underwater robots by physically combining multiple units. This architecture eliminates centralized wiring and enables rapid structural reconfiguration.

By combining multiple modules, users can construct various underwater robot configurations depending on the intended motion and experimental objective.

\subsection{Waterproof Self-Contained Module Design}

\begin{figure}[t]
  \centering  \includegraphics[width=\columnwidth]{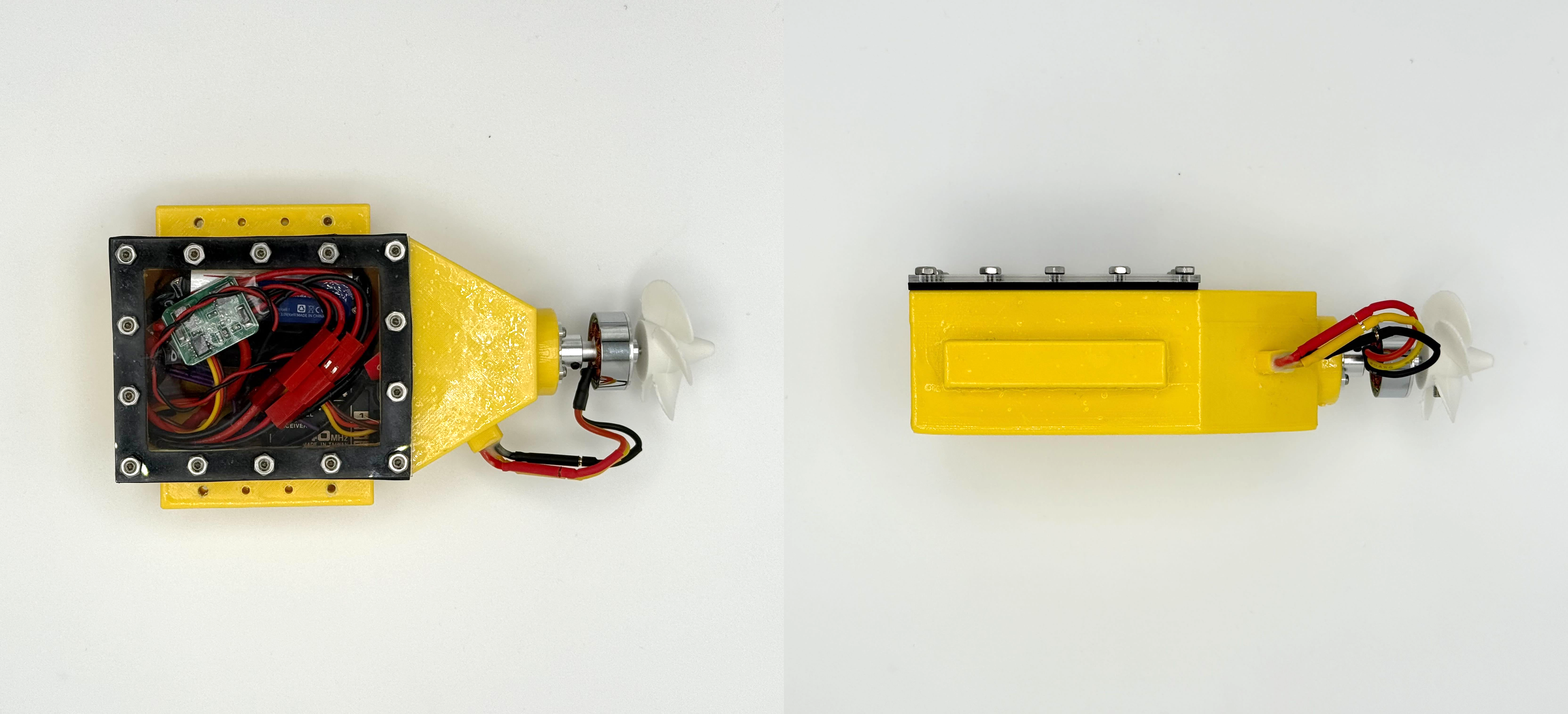}
  \caption{Self-contained propulsion module. The waterproof housing integrates battery, wireless receiver, and motor driver, while the brushless motor is mounted externally.}
  \label{fig:module}
\end{figure}

Each propulsion module measures approximately 170\,mm $\times$ 80\,mm $\times$ 50\,mm (length $\times$ width $\times$ height) and weighs approximately 257\,g (Fig.~\ref{fig:module}). The module consists of a sealed waterproof housing that protects internal electronics from water exposure. A rechargeable battery, a 40\,MHz wireless receiver, and a motor driver circuit are integrated within the enclosure, forming a self-contained propulsion unit. The brushless motor is mounted externally, while the internal circuitry remains fully sealed.

To improve usability and reduce mechanical wear, power switching is implemented using a magnetic mechanism. This allows the module to be turned on and off without opening the housing, thereby preserving waterproof integrity during repeated use.

\subsection{Wireless Distributed Control}

\begin{figure}[t]
  \centering
  \includegraphics[width=\columnwidth]{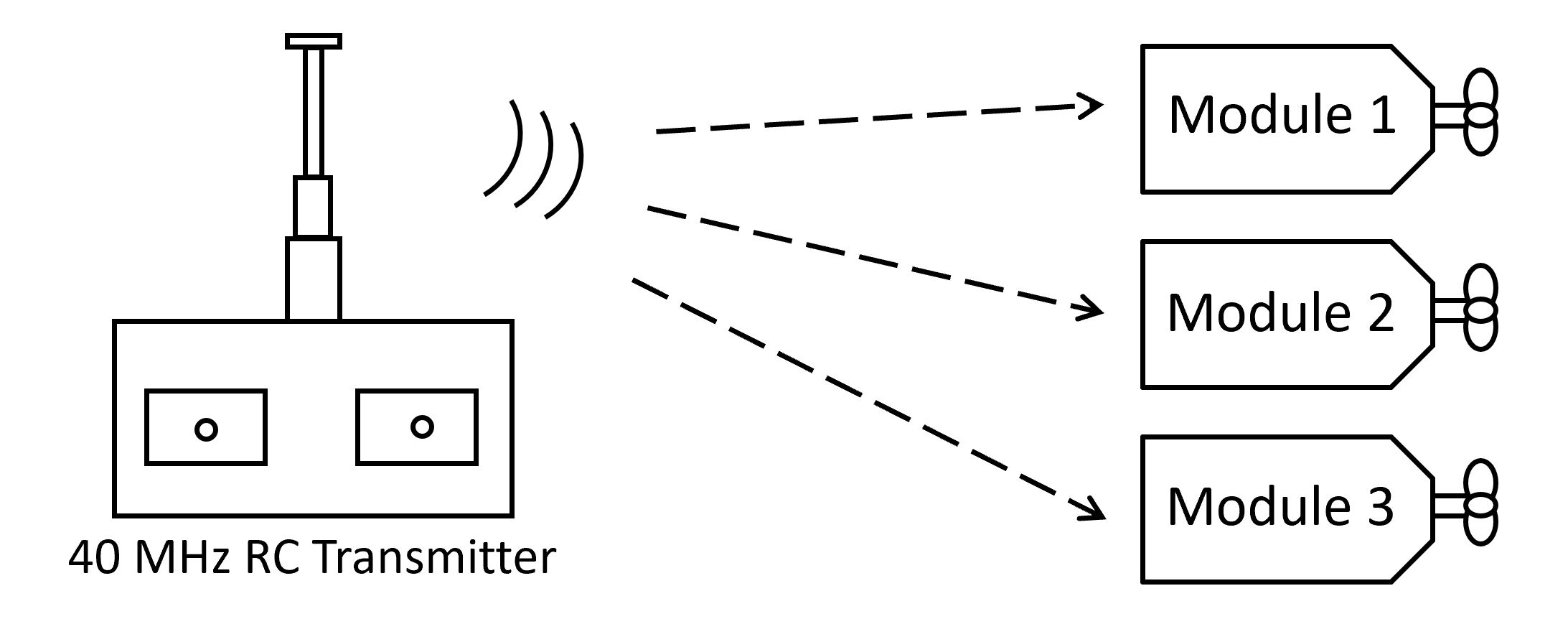}
  \caption{Wireless control architecture of MarineCraft. Each module contains an independent 40\,MHz receiver and is directly controlled by a radio transmitter without centralized wiring.}
  \label{fig:system}
\end{figure}

Control commands are transmitted from a single Futaba transmitter to each module independently (Fig.~\ref{fig:system}). By assigning each module's receiver to an independent channel on the transmitter, a single operator can control multiple propulsion units simultaneously. This enables complex maneuvers, such as differential steering, without signal interference. Since each unit is self-powered, this distributed approach eliminates the need for any physical wiring during reconfiguration.

\subsection{Mechanical Connection and Reconfiguration}

Each module includes external connection tabs that enable mechanical attachment through dedicated joints. These joints support various geometric arrangements, including linear and parallel configurations.

The mechanical interface is designed to allow repeated assembly and disassembly while maintaining structural stability during underwater operation.

\subsection{Buoyancy and Stability Adjustment}

Underwater stability can be tuned using attachable ballast weights and buoyancy elements. After assembly, each configuration is tested for waterproof integrity and floated on the water surface. If the robot does not maintain the intended orientation, ballast weights are added to modify mass distribution. Neutral buoyancy is achieved by balancing total weight and buoyant force using attachable foam components.

These external adjustment mechanisms enable iterative tuning of underwater behavior without modifying the internal electronics.

\section{Implementation}

The module housing was fabricated using fused deposition modeling (FDM) 3D printing. Polylactic acid (PLA) was selected as the printing material due to its availability, ease of processing, and sufficient mechanical rigidity for shallow-water applications. After printing, support materials were carefully removed to ensure clean internal surfaces and proper sealing interfaces.

To achieve waterproof performance, resin coating was applied to both the internal and external surfaces of the housing. The resin was manually applied and cured to create a continuous sealing layer. Particular attention was paid to edges, joint boundaries, and cable entry points to minimize leakage risk.

A brushless DC motor was selected as the propulsion unit and mounted externally on the housing to avoid internal water exposure. Electrical wiring connects the motor to the internal motor driver circuit, and all cable interfaces are sealed to prevent water ingress. Internal components, including the battery, wireless receiver, and motor driver, are arranged within the sealed compartment to maintain compact integration.

To allow internal inspection and maintenance, part of the housing is covered with a transparent acrylic plate. This plate is secured using bolts and nuts, enabling easy removal while maintaining sealing pressure during operation. A rubber sheet gasket is inserted between the acrylic cover and the housing to improve waterproof sealing at the joint interface.

Waterproof verification was conducted using a pressure-based leak test. A syringe was attached to a sealed port on the acrylic cover, and internal air pressure was increased while the module was submerged in water. The presence of air bubbles indicated leakage points. When leakage was detected, additional resin was applied to the identified areas, and the sealing test was repeated until no air escape was observed.

\section{Experimental Evaluation}

\begin{figure}[t]
\centering
\includegraphics[width=\columnwidth]{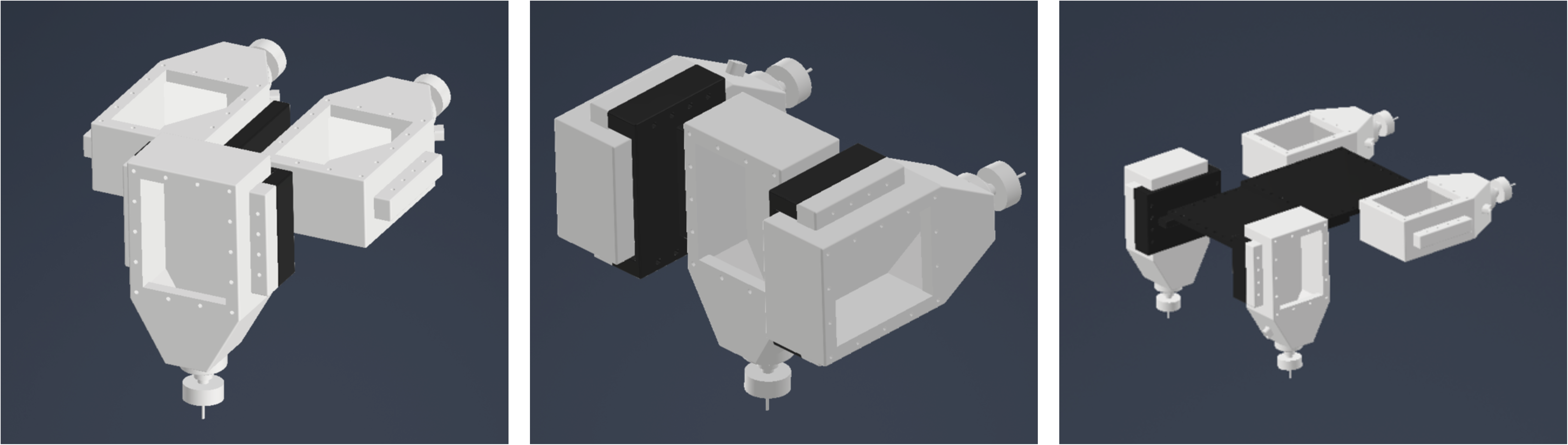}
\caption{Three distinct CAD-designed underwater robot configurations assembled from MarineCraft modules. The configurations illustrate the structural diversity enabled by modular mechanical reconfiguration without electrical rewiring.}
\label{fig:cad}
\end{figure}

\begin{figure}[t]
  \centering
  \includegraphics[width=\columnwidth]{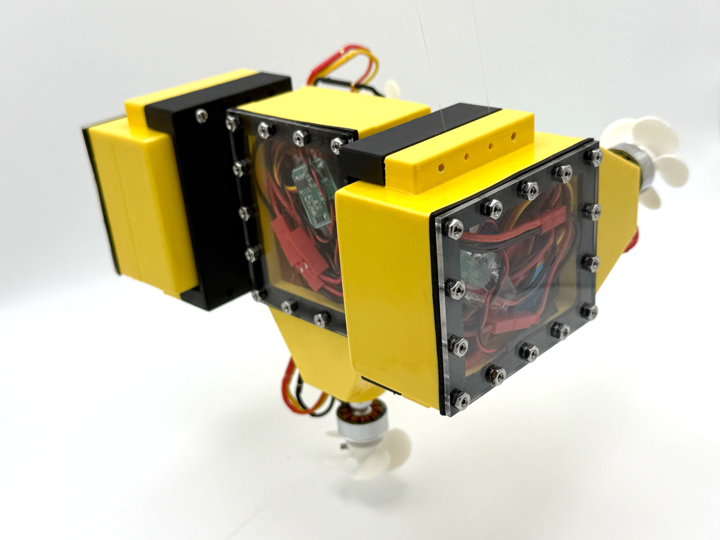}
  \caption{Representative underwater robot configurations constructed using MarineCraft modules. Diverse geometries were realized through mechanical assembly without electrical rewiring.}
  \label{fig:differenttype}
\end{figure}

\begin{figure}[t]
  \centering
  \includegraphics[width=\columnwidth]{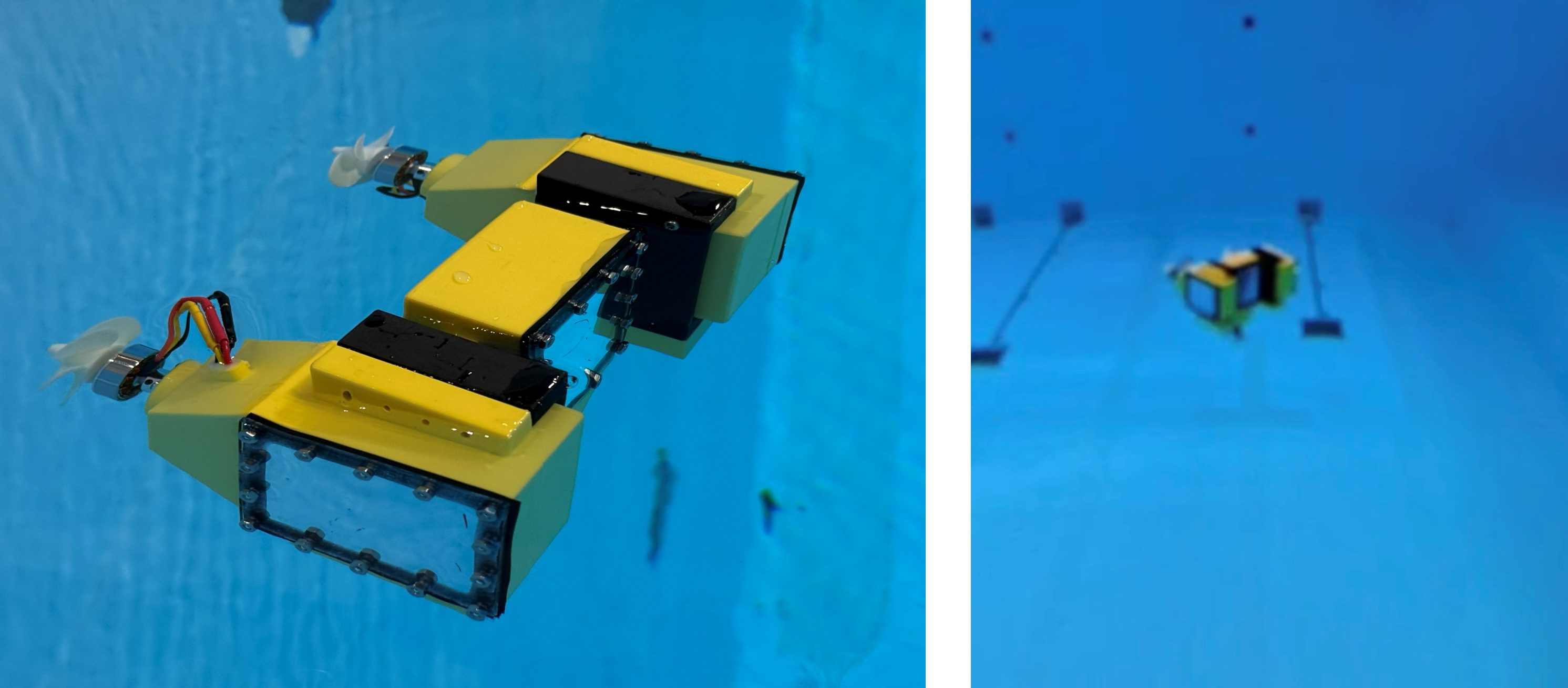}
  \caption{Underwater operation test conducted in a swimming pool environment. The robot maintained waterproof integrity and achieved controlled propulsion at depths of up to approximately 2.5 meters.}
  \label{fig:underwaterexp}
\end{figure}

To evaluate the effectiveness of MarineCraft as a modular underwater robot construction toolkit, we conducted experiments focusing on structural reconfigurability and functional underwater operation.

\subsection{Structural Reconfigurability}

We first examined whether the proposed module enables the construction of diverse underwater robot geometries without modifying internal electronics or wiring. The propulsion modules were fabricated using a 3D printer and mechanically combined through dedicated joints and connectors. By rearranging module positions and orientations, multiple robot configurations with distinct geometric layouts were constructed using identical propulsion units.

Figure~\ref{fig:cad} presents representative CAD-designed underwater robot configurations assembled from the same self-contained modules. Among these, the configurations shown in Fig.~\ref{fig:fig1} and Fig.~\ref{fig:differenttype} were physically fabricated to validate the design-to-implementation consistency of the modular system. All configurations were realized through mechanical reconfiguration alone, without electrical rewiring or internal modification.

These results demonstrate that the proposed modular architecture supports rapid structural reconfiguration and enables diverse underwater robot geometries to be realized using accessible 3D printing tools. The separation between structural layout and functional propulsion units allows scalable morphological exploration while maintaining consistent system behavior across configurations.

\subsection{Underwater Operation Test}

To validate underwater functionality, the fabricated robot configurations were tested in a swimming pool environment. Each configuration was fully submerged and operated using wireless control. During the experiments, no internal water leakage was observed in any module (Fig.~\ref{fig:underwaterexp}).

Each configuration achieved approximately five minutes of continuous underwater movement, including directional maneuvering. The robots demonstrated stable operation at depths up to 2.5 meters, which corresponded to the maximum depth of the testing facility.

These results confirm that the waterproof sealing method and integrated self-contained module design provide reliable underwater operation while preserving structural reconfigurability. The experiments demonstrate that MarineCraft enables both flexible morphological exploration and practical underwater deployment in shallow-water environments.

\section{Discussion and Future Work}

While MarineCraft demonstrates structural reconfigurability and reliable short-term underwater operation, several areas for improvement remain to be addressed in future work.

First, although the current system allows users to intuitively learn about buoyancy and stability through hands-on trial-and-error—a core educational strength of MarineCraft—this process can be further optimized. We plan to develop a design-support software tool that visualizes mass distribution and buoyancy based on module arrangements. Rather than replacing the physical experience, this tool will complement the "build-and-test" workflow by providing predictive insights, helping users navigate complex design spaces more effectively.

Second, the current evaluation focused on functional validation rather than rigorous quantitative analysis. Future studies will involve detailed measurements of propulsion efficiency, energy consumption, and maneuverability under hydrodynamic disturbances. Furthermore, long-term durability in saltwater and structural integrity at depths beyond 2.5 meters must be investigated to transition from shallow-water prototyping to extended field deployment.

Finally, expanding the module library to include sensing units and autonomous control components represents a promising direction. These developments will transform MarineCraft from a manual construction toolkit into a comprehensive platform for both educational exploration and advanced underwater robotics research.

\section{Conclusion}

This paper presented MarineCraft, a modular underwater robot construction platform designed to lower the barrier to hands-on fabrication in educational contexts. The proposed system consists of waterproof, self-contained propulsion modules integrating power, wireless communication, and actuation within compact units. The modular architecture enables structural reconfiguration without centralized wiring, allowing users to assemble diverse underwater robot geometries using accessible 3D printing tools.

Experimental evaluation demonstrated that multiple robot configurations can be constructed through mechanical assembly alone and successfully operated underwater. The same propulsion modules were reused across different robot geometries without rewiring or reopening the waterproof enclosures, demonstrating that the proposed architecture supports iterative structural exploration with minimal preparation effort. The robots achieved continuous movement for approximately five minutes per configuration and descended to depths of up to approximately 2.5 meters without internal water leakage.

These results indicate that MarineCraft provides a practical and physically grounded platform for exploring underwater robot morphology while maintaining reliable operation in shallow-water environments. By separating structural reconfiguration from waterproof electrical implementation, MarineCraft simplifies the iterative design process and supports rapid prototyping of underwater robot configurations. Future work includes quantitative performance analysis, comparisons with conventional construction workflows, long-duration endurance testing, and the development of additional modular components to further expand structural and functional capabilities.

\section*{Acknowledgments}

This paper was prepared with the assistance of generative AI tools (e.g., for language refinement and drafting support). All content has been reviewed and verified by the authors.

% This section is left blank for anonymous review.

%

%-----------------------------------------------------------------------

%%%%%%%%%%%%%%%% BIBLIOGRAPHY IN THE LaTeX file !!!!! %%%%%%%%%%%%%%%%%%%%%%

\end{document}